\pdfoutput=1
\documentclass[11pt]{article}

\usepackage[final]{acl}

\usepackage{times}
\usepackage{latexsym}
\usepackage{multirow}
\usepackage{graphicx}

\usepackage[T1]{fontenc}
\usepackage[utf8]{inputenc}

\usepackage{microtype}

\usepackage{inconsolata}

\title{MediFact at MEDIQA-M3G 2024: Medical Question Answering in Dermatology with Multimodal Learning}

\author{Nadia Saeed \\
  Computational Biology Research Lab \\ Department of Computer Science \\
  National University of
Computer and Emerging Sciences (NUCES-FAST)\\ Islamabad, Pakistan \\
  \texttt{i181606@nu.edu.pk} \\}

\begin{document}
\maketitle
\begin{abstract}
The MEDIQA-M3G 2024 challenge necessitates novel solutions for Multilingual \& Multimodal Medical Answer Generation in dermatology \cite{mediqa-m3g-task}. This paper addresses the limitations of traditional methods by proposing a weakly supervised learning approach for open-ended medical question-answering (QA). Our system leverages readily available MEDIQA-M3G images via a VGG16-CNN-SVM model, enabling multilingual (English, Chinese, Spanish) learning of informative skin condition representations. Using pre-trained QA models, we further bridge the gap between visual and textual information through multimodal fusion. This approach tackles complex, open-ended questions even without predefined answer choices. We empower the generation of comprehensive answers by feeding the ViT-CLIP model with multiple responses alongside images. This work advances medical QA research, paving the way for clinical decision support systems and ultimately improving healthcare delivery. \footnote{Fine-tuned models and Code avaliable: \url{https://github.com/NadiaSaeed/MediFact-M3G-MEDIQA-2024}}
\end{abstract}

\section{Introduction}
Dermatological telemedicine consultations, while offering a promising solution for remote diagnosis and treatment, face hurdles due to limitations in capturing subtle visual details and the inability to physically examine lesions. This can lead to miscommunication, such as difficulties in describing the texture or progression of lesions, which can hinder the development of effective treatment plans \cite{elsner2020teledermatology, hwang2024review, mehraeen2023telemedicine}. However, recent advancements in image-text learning, like Vision Transformer (ViT) for image captioning and Contrastive Language-Image Pre-Training (CLIP) for aligning text and image representations, offer promising avenues to bridge this gap \cite{yin2022vit,li2021supervision}.

\begin{figure*}[h]
  \includegraphics[width=\textwidth,height=14cm]{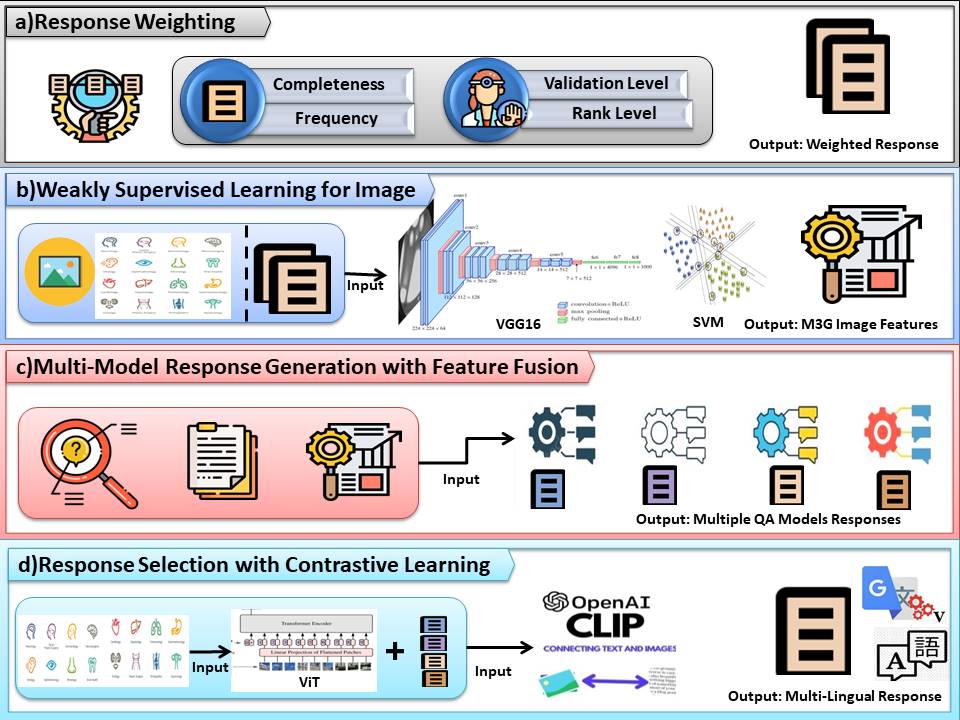}
  \caption{MediFact-M3G Framework: From Uncertain Data to Informed Answers}
  \label{fig:02}
\end{figure*}

Existing approaches to teledermatology consultations have limitations. Traditional consumer health question-answering systems primarily focus on textual data, neglecting the valuable information within visual details \cite{abacha2019overview}. This limits their ability to understand the nuances of skin conditions often best captured visually. Visual question-answering efforts have mainly targeted radiology images, overlooking the crucial context provided by clinical text \cite{abacha2019vqa}. While recent advancements in deep learning have shown promise in lesion classification for dermatology \cite{li2022artificial}, these approaches often focus on specific image types and cannot integrate textual information, essential for a holistic understanding of a patient's condition. While some research explores combining clinical text and images for specific dermatology tasks, such as melanoma risk assessment, they haven't addressed open-ended question answering \cite{groh2022towards,lin2023medical}. 

This research tackles these limitations by introducing a novel framework for multilingual and multimodal query response generation in clinical dermatology. Our system leverages the power of multimodal fusion, which combines information from different sources. In this case, the sources are textual and visual: textual clinical context and user queries in multiple languages, along with user-uploaded images. This work introduces Medifact-M3G, a framework for tackling uncertainties in medical question answering for dermatology shown in Figure \ref{fig:02}. Medifact-M3G prepares the data and assigns weights to potential answers, considering their relevance and trustworthiness (Section a). It then uses a powerful image analysis tool to extract key features from skin condition images (Section b). By combining these features with text analysis, Medifact-M3G leverages multiple powerful models to generate informative answers to medical questions (Sections c and d). This framework has the potential to improve the accuracy and reliability of AI-powered diagnosis systems in telemedicine, ultimately assisting healthcare professionals in providing better diagnoses and treatment plans. This research addresses the following key questions:
1) Can feature fusion from weakly supervised learning techniques effectively support open-ended medical question answering in dermatology?
2) Can a Medifact-M3G fine-tuned model trained solely on the MEDIQA-M3G training dataset adequately capture similarities and relatedness for unseen samples?
3) How can contrastive learning be seamlessly integrated with Medifact-M3G to quantify uncertainty in response generation for ambiguous queries and limited content information?
\section{Methodology} 
Our response generation system for the MEDIQA-M3G 2024 task tackles the challenge of limited labeled data while aiming to generate informative responses to user queries about dermatological conditions \cite{mediqa-m3g-task}. This methodology leverages several key steps, as illustrated in the accompanying MediFact-M3G framework shown in Figure \ref{fig:02}.
\begin{figure*}[h]
  \includegraphics[width=\textwidth,height=16cm]{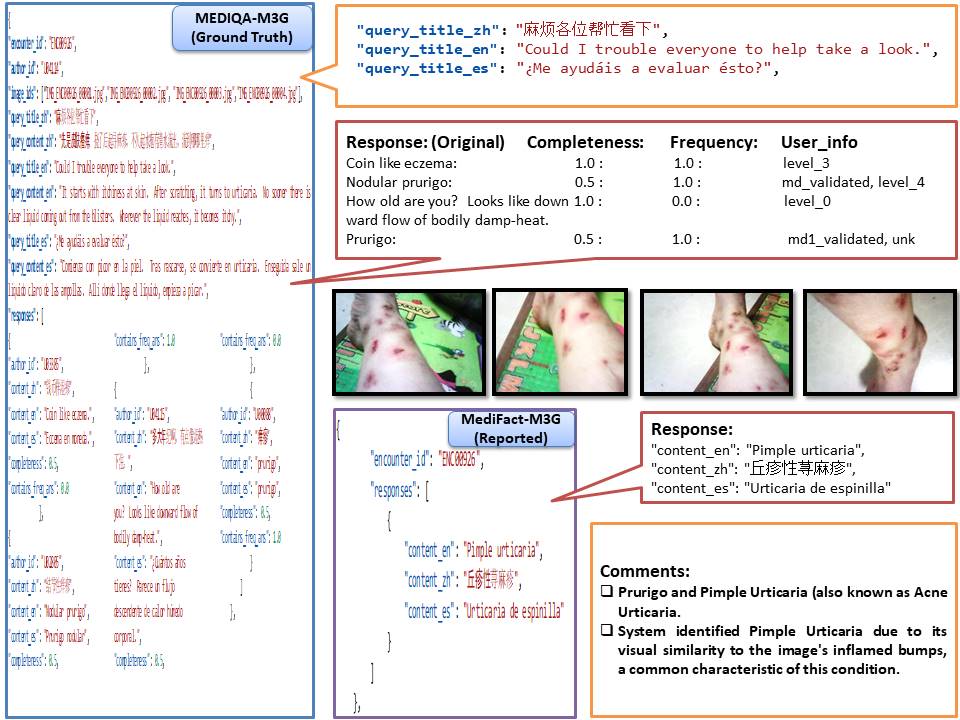}
  \caption{Example of Original Text "MEDIQA-M3G" and System Output "MediFact-M3G}
  \label{fig:03}
\end{figure*}
\subsection{Data Preprocessing and Response Weighting}

We begin by ensuring the quality of the raw data through techniques like handling missing values, text cleaning, and formatting consistency.  This establishes a clean and consistent foundation for subsequent model training.

Next, a weighting function assigns scores to each response based on the author's expertise (e.g., medical doctor) and response completeness. This guides the model to prioritize learning from the most effective responses during training, ultimately improving the quality of the generated responses.

\subsection{Weakly Supervised Learning for Image Representation: Addressing Data Limitations}

While large, labeled datasets are ideal for training robust response generation models in dermatology, ethical considerations, and data access limitations often restrict their availability. To address this challenge, we employed a weakly supervised learning approach that leverages the available data effectively.

Our approach utilizes a pre-trained Convolutional Neural Network (CNN), specifically VGG16, to extract high-level features from the dermatological images. These features capture the visual characteristics relevant to diagnosis \cite{desai2021hybrid}. We then use a Support Vector Machine (SVM) classifier to learn the relationship between the extracted image features and the high-quality textual responses associated with labeled image-response pairs. The SVM essentially learns to map images to their most relevant textual descriptions \cite{chandra2021survey}.

This weakly supervised approach allows us to overcome limitations in labeled data. The SVM generalizes the learned relationship between labeled image-response pairs to unlabeled images. By incorporating the information gleaned from the textual responses, this process enriches the image representations learned by the VGG16 model, even without explicit labels for each unlabeled image. These enriched image representations (English, Chinese, and Spanish languages) capture the semantic meaning associated with the images, providing valuable information for the response generation model during training. Additionally, for comparison purposes, we evaluated the performance of Inception and ResNet models in place of VGG16 to determine the most effective CNN architecture for this task \cite{zheng2021mutual,zhou2021covid}.

\subsection{Multi-Model Response Generation with Feature Fusion}

This step focuses on generating responses to user queries. We employ a multi-model approach that combines pre-trained question-answering (QA) models with the image representation learned from the weakly supervised approach described in Section 2 \cite{cortiz2022exploring}. Due to limitations in the performance and availability of non-English language models, this step focuses on English responses.

A comprehensive feature vector for each query-response pair is created by combining the following elements:
\begin{itemize}
\item The user's query itself.
\item Relevant textual content (e.g., patient demographics).
\item The image representation learned from the weakly supervised approach (Section 2).
\end{itemize}
We utilize two pre-trained English models:

\subsubsection{Extractive QA Model} This model retrieves relevant answer passages from a text corpus (potentially including high-quality responses) that directly address the user's query \cite{guo2023msq,clark2020electra, he2021debertav3}.
\subsubsection{Abstractive QA Model} This model goes beyond retrieval and generates a new, comprehensive response. It incorporates information from various sources (textual features, extracted passages) and potential reasons over the information to provide a more informative answer \cite{lewis2019bart}.

This multi-model approach offers the advantage of combining factual grounding from the extractive model with flexible response generation from the abstractive model, while also incorporating visual information through the image features. This ultimately leads to more accurate and informative responses within the teledermatology domain.

\subsection{Response Selection with Contrastive Learning}
Selecting the most informative response for a query-image pair, especially in non-English settings, requires a robust approach. We leverage CLIP, a contrastive learning model adept at learning relationships between image and text embeddings \cite{li2021supervision}. CLIP utilizes a Vision Transformer (ViT) (Section 2) to extract high-dimensional image features and a separate text encoder for potential responses \cite{yin2022vit}.
We employ CLIP in two key settings: First, CLIP receives the ViT-extracted image embedding and multiple response lists (English, Spanish, Chinese). It calculates the cosine similarity between each response embedding (in a specific language) and the image embedding. The response with the highest similarity (closest semantic relationship) is chosen for that language.
Second, CLIP focuses on the relationship between the image and English responses from pre-trained QA models (Section 3). It assesses the cosine similarity between the image embedding and the selected English response embedding. Google Translate then converts this English response to Spanish and Chinese for user convenience, acknowledging potential translation inaccuracies \cite{taira2021pragmatic}. 
\section {Experimental Setup and Results}
We evaluated our model's capability in addressing the problem of clinical dermatology multimodal query response generation. This evaluation was conducted within the Shared Task of MEDIQA-M3G 2024, which focuses on multilingual and multimodal medical answer generation \cite{mediqa-m3g-task}. As illustrated in Figure \ref{fig:03}, each sample in the task comprised k medical images related to dermatological conditions, a textual query describing the user's skin concern, and its content. Additionally, the ground truth for each sample included multiple possible responses with corresponding scores. Leveraging the framework outlined in Figure \ref{fig:02}, our Medifact-M3G model was employed to generate answers in three languages for each sample.

\subsection{Dataset}
The MEDIQA-M3G dataset is divided into training (842 instances), validation (56 instances), and test (100 instances) sets, with each set available in Chinese, English, and Spanish versions \cite{mediqa-m3g-dataset}. While non-English training sets are machine-translated, validation and test sets are human-translated for accuracy. Each instance is represented as a JSON object containing a unique encounter ID, a list of image IDs, the query title and content in the specific language, and author information from a separate CSV file. Participants are expected to generate responses in JSON format, including a unique encounter ID and a list of generated responses for the specified languages. Participation in all language evaluations is optional, with empty strings allowed for non-participating languages.

\subsection{Evaluation Metrics}
Our system's performance was evaluated using official available evaluation program of MEDIQA-M3G \footnote{MEDIQA-M3G evaluation code: \url{https://github.com/wyim/MEDIQA-M3G-2024/tree/main}}. metrics commonly employed in Natural Language Generation (NLG) tasks. DeltaBLEU and BERTScore were chosen for this assessment [cite]. DeltaBLEU measures the similarity between a generated response and reference responses by considering n-gram (sequence of n words) overlap but weighs these n-grams based on human judgment. BERTScore, on the other hand, focuses on the semantic similarity between the generated response and the references, taking the maximum score from any available reference response. The evaluation script processed instances across three languages (English, Spanish, and Chinese).

\subsection{Result}

In this study, we evaluated our approach using the Mediqa-M3G framework, employing three different feature extraction models while maintaining consistency in other aspects of the setup. These models included SVMs with default sklearn settings and pre-trained CNN architectures like from the Keras library. The evaluation results are summarized in Table \ref{table01}.

\begin{table*}[h]
\centering
\begin{tabular}{|ll|lll|lll|}
\hline
\multicolumn{2}{|l|}{}                                                  & \multicolumn{3}{l|}{\textbf{Deltableu}}                                                    & \multicolumn{3}{l|}{\textbf{BERT\_Score}}                                                                \\ \cline{3-8} 
\multicolumn{2}{|l|}{\multirow{-2}{*}{\textbf{Model}}}                  & \multicolumn{1}{l|}{en}             & \multicolumn{1}{l|}{zh}             & es             & \multicolumn{1}{l|}{en}             & \multicolumn{1}{l|}{zh}                           & es             \\ \hline
\multicolumn{1}{|l|}{}                               & VGG16-Individual & \multicolumn{1}{l|}{0.588}          & \multicolumn{1}{l|}{4.503}          & \textbf{0.918} & \multicolumn{1}{l|}{0.837}          & \multicolumn{1}{l|}{\textbf{0.771}}               & 0.804          \\ \cline{2-8} 
\multicolumn{1}{|l|}{\multirow{-2}{*}{MediFact-M3G}} & VGG16-Translated & \multicolumn{1}{l|}{\textbf{0.717}} & \multicolumn{1}{l|}{\textbf{4.503}} & 0.823          & \multicolumn{1}{l|}{\textbf{0.842}} & \multicolumn{1}{l|}{{0.763}} & \textbf{0.809} \\ \hline
\end{tabular}
\caption{Scores for Response Generation Approaches on MEDIQA-M3G Testing Dataset (submitted at the competition)}
\label{table01}
\end{table*}

Table \ref{table01} displays the evaluation results for two setups of the MediFact-M3G framework. In the first setup, denoted as VGG16-Individual, separate VGG16 models were trained for each language, yielding individual scores for each language. In the second setup, the best-performing VGG16 model output, which was observed to be the Chinese language model, was utilized to translate responses into English and Spanish languages following the MediFact-M3G framework. While the translated version of MediFact-M3G showed slight improvement in BERT\_Score, the Deltableu score performed better in the individual setup for Spanish language responses.

Additionally, it's worth noting our performance in the MEDIQA-M3G 2024 shared task, where we achieved 7th rank in English language response generation, and 3rd rank in Chinese and Spanish language response generation, out of a total of 75 participants. These rankings underscore the effectiveness of our approach across different languages and its competitiveness in challenging benchmark tasks \cite{mediqa-m3g-task}.
\subsection{Discussion}

The results presented here were obtained after rigorous testing in a challenging setting, providing insights into the performance of different feature extraction models within the MediFact-M3G framework. VGG16-Translated demonstrated significant improvements over VGG16-Individual, underscoring the effectiveness of data translation in enhancing translation quality. The evaluation results are summarized in Table \ref{table_squeezenet}.

\begin{table*}[h]
\centering
\begin{tabular}{|cl|lll|lll|}
\hline
\multicolumn{2}{|l|}{}                                                  & \multicolumn{3}{l|}{\textbf{Deltableu}}                                                                                                       & \multicolumn{3}{l|}{\textbf{BERT\_Score}}                                                                                                     \\ \cline{3-8} 
\multicolumn{2}{|l|}{\multirow{-2}{*}{\textbf{Model}}}                  & \multicolumn{1}{l|}{en}                           & \multicolumn{1}{l|}{zh}                           & es                                    & \multicolumn{1}{l|}{en}                                    & \multicolumn{1}{l|}{zh}                           & es                           \\ \hline
\multicolumn{1}{|c|}{}                               & VGG16-Individual & \multicolumn{1}{l|}{{\color[HTML]{000000} 0.588}} & \multicolumn{1}{l|}{{\color[HTML]{000000} 4.503}} & {\color[HTML]{000000} \textbf{0.918}} & \multicolumn{1}{l|}{{\color[HTML]{000000} \textbf{0.845}}} & \multicolumn{1}{l|}{{\color[HTML]{000000} 0.763}} & {\color[HTML]{000000} 0.806} \\ \cline{2-8} 
\multicolumn{1}{|c|}{}                               & VGG16-Translated & \multicolumn{1}{l|}{0.717}                        & \multicolumn{1}{l|}{4.503}                        & 0.823                                 & \multicolumn{1}{l|}{0.842}                                 & \multicolumn{1}{l|}{0.763}                        & \textbf{0.809}               \\ \cline{2-8} 
\multicolumn{1}{|c|}{}                               & ResNet           & \multicolumn{1}{l|}{0.565}                        & \multicolumn{1}{l|}{\textbf{6.457}}               & 0.542                                 & \multicolumn{1}{l|}{0.837}                                 & \multicolumn{1}{l|}{\textbf{0.771}}               & 0.804                        \\ \cline{2-8} 
\multicolumn{1}{|c|}{\multirow{-4}{*}{MediFact-M3G}} & SqueezNet        & \multicolumn{1}{l|}{\textbf{0.744}}               & \multicolumn{1}{l|}{2.125}                        & 0.641                                 & \multicolumn{1}{l|}{0.841}                                 & \multicolumn{1}{l|}{0.702}                        & 0.808                        \\ \hline
\end{tabular}
\caption{{Scores for Response Generation Approaches on MEDIQA-M3G Testing Dataset (after the competition)}}
\label{table_squeezenet}
\end{table*}

After replacing the VGG16 models with ResNet and SqueezNet in MediFact-M3G framework, we obtained the following evaluation results as shown in Table \ref{table_squeezenet}. SqueezNet demonstrated exceptional proficiency in Chinese translations, achieving the highest Deltableu scores across all languages. On the other hand, although ResNet exhibited slightly lower Deltableu scores, its competitive performance across all languages highlights its versatility in handling various translation tasks. These findings underscore the critical role of selecting appropriate feature extraction models tailored to specific language requirements and task objectives, ultimately enhancing the effectiveness of the MediFact-M3G framework in addressing medical query challenges.
\section{Future Work}
In the future, we plan to conduct further experiments to explore the robustness and scalability of our approach across larger and more diverse datasets. Additionally, we aim to investigate the integration of domain-specific ontologies and medical terminologies to enhance the semantic understanding and accuracy of our system. Furthermore, we are interested in exploring novel techniques for handling multi-turn dialogue scenarios, allowing our system to engage in more natural and interactive conversations with users. Additionally, we plan to collaborate with medical professionals to validate the clinical relevance and effectiveness of our approach in real-world healthcare settings. By addressing these challenges, we hope to continue advancing the field of medical question-answering and contribute to the development of more practical and clinically useful systems.

\bibliography{acl_latex}

\begin{thebibliography}{22}
\expandafter\ifx\csname natexlab\endcsname\relax\def\natexlab#1{#1}\fi

\bibitem[{Abacha et~al.(2019{\natexlab{a}})Abacha, Hasan, Datla, Liu, Demner-Fushman, and M{\"u}ller}]{abacha2019vqa}
Asma~Ben Abacha, Sadid~A Hasan, Vivek~V Datla, Joey Liu, Dina Demner-Fushman, and Henning M{\"u}ller. 2019{\natexlab{a}}.
\newblock Vqa-med: Overview of the medical visual question answering task at imageclef 2019.
\newblock \emph{CLEF (working notes)}, 2(6).

\bibitem[{Abacha et~al.(2019{\natexlab{b}})Abacha, Shivade, and Demner-Fushman}]{abacha2019overview}
Asma~Ben Abacha, Chaitanya Shivade, and Dina Demner-Fushman. 2019{\natexlab{b}}.
\newblock Overview of the mediqa 2019 shared task on textual inference, question entailment and question answering.
\newblock In \emph{Proceedings of the 18th BioNLP Workshop and Shared Task}, pages 370--379.

\bibitem[{Chandra and Bedi(2021)}]{chandra2021survey}
Mayank~Arya Chandra and SS~Bedi. 2021.
\newblock Survey on svm and their application in image classification.
\newblock \emph{International Journal of Information Technology}, 13(5):1--11.

\bibitem[{Clark et~al.(2020)Clark, Luong, Le, and Manning}]{clark2020electra}
Kevin Clark, Minh-Thang Luong, Quoc~V. Le, and Christopher~D. Manning. 2020.
\newblock \href {http://arxiv.org/abs/2003.10555} {Electra: Pre-training text encoders as discriminators rather than generators}.

\bibitem[{Cortiz(2022)}]{cortiz2022exploring}
Diogo Cortiz. 2022.
\newblock Exploring transformers models for emotion recognition: A comparision of bert, distilbert, roberta, xlnet and electra.
\newblock In \emph{Proceedings of the 2022 3rd International Conference on Control, Robotics and Intelligent System}, pages 230--234.

\bibitem[{Desai et~al.(2021)Desai, Pujari, Sujatha, Kamble, and Kambli}]{desai2021hybrid}
Padmashree Desai, Jagadeesh Pujari, C~Sujatha, Arinjay Kamble, and Anusha Kambli. 2021.
\newblock Hybrid approach for content-based image retrieval using vgg16 layered architecture and svm: an application of deep learning.
\newblock \emph{SN Computer Science}, 2(3):170.

\bibitem[{Elsner(2020)}]{elsner2020teledermatology}
Peter Elsner. 2020.
\newblock Teledermatology in the times of covid-19--a systematic review.
\newblock \emph{JDDG: Journal Der Deutschen Dermatologischen Gesellschaft}, 18(8):841--845.

\bibitem[{Groh et~al.(2022)Groh, Harris, Daneshjou, Badri, and Koochek}]{groh2022towards}
Matthew Groh, Caleb Harris, Roxana Daneshjou, Omar Badri, and Arash Koochek. 2022.
\newblock Towards transparency in dermatology image datasets with skin tone annotations by experts, crowds, and an algorithm.
\newblock \emph{Proceedings of the ACM on Human-Computer Interaction}, 6(CSCW2):1--26.

\bibitem[{Guo et~al.(2023)Guo, Guo, Dougherty, and Jin}]{guo2023msq}
Muzhe Guo, Muhao Guo, Edward~T Dougherty, and Fang Jin. 2023.
\newblock Msq-biobert: Ambiguity resolution to enhance biobert medical question-answering.
\newblock In \emph{Proceedings of the ACM Web Conference 2023}, pages 4020--4028.

\bibitem[{He et~al.(2021)He, Gao, and Chen}]{he2021debertav3}
Pengcheng He, Jianfeng Gao, and Weizhu Chen. 2021.
\newblock Debertav3: Improving deberta using electra-style pre-training with gradient-disentangled embedding sharing.
\newblock \emph{arXiv preprint arXiv:2111.09543}.

\bibitem[{Hwang et~al.(2024)Hwang, Del~Toro, Han, Oh, Tejasvi, and Lipner}]{hwang2024review}
Jonathan~K Hwang, Natalia~Pelet Del~Toro, George Han, Dennis~H Oh, Trilokraj Tejasvi, and Shari~R Lipner. 2024.
\newblock Review of teledermatology: lessons learned from the covid-19 pandemic.
\newblock \emph{American Journal of Clinical Dermatology}, 25(1):5--14.

\bibitem[{Lewis et~al.(2019)Lewis, Liu, Goyal, Ghazvininejad, Mohamed, Levy, Stoyanov, and Zettlemoyer}]{lewis2019bart}
Mike Lewis, Yinhan Liu, Naman Goyal, Marjan Ghazvininejad, Abdelrahman Mohamed, Omer Levy, Ves Stoyanov, and Luke Zettlemoyer. 2019.
\newblock \href {http://arxiv.org/abs/1910.13461} {Bart: Denoising sequence-to-sequence pre-training for natural language generation, translation, and comprehension}.

\bibitem[{Li et~al.(2021)Li, Liang, Zhao, Cui, Ouyang, Shao, Yu, and Yan}]{li2021supervision}
Yangguang Li, Feng Liang, Lichen Zhao, Yufeng Cui, Wanli Ouyang, Jing Shao, Fengwei Yu, and Junjie Yan. 2021.
\newblock Supervision exists everywhere: A data efficient contrastive language-image pre-training paradigm.
\newblock In \emph{International Conference on Learning Representations}.

\bibitem[{Li et~al.(2022)Li, Koban, Schenck, Giunta, Li, and Sun}]{li2022artificial}
Zhouxiao Li, Konstantin~Christoph Koban, Thilo~Ludwig Schenck, Riccardo~Enzo Giunta, Qingfeng Li, and Yangbai Sun. 2022.
\newblock Artificial intelligence in dermatology image analysis: current developments and future trends.
\newblock \emph{Journal of clinical medicine}, 11(22):6826.

\bibitem[{Lin et~al.(2023)Lin, Zhang, Tao, Shi, Haffari, Wu, He, and Ge}]{lin2023medical}
Zhihong Lin, Donghao Zhang, Qingyi Tao, Danli Shi, Gholamreza Haffari, Qi~Wu, Mingguang He, and Zongyuan Ge. 2023.
\newblock Medical visual question answering: A survey.
\newblock \emph{Artificial Intelligence in Medicine}, page 102611.

\bibitem[{Mehraeen et~al.(2023)Mehraeen, SeyedAlinaghi, Heydari, Karimi, Mahdavi, Mashoufi, Sarmad, Mirghaderi, Shamsabadi, Qaderi et~al.}]{mehraeen2023telemedicine}
Esmaeil Mehraeen, SeyedAhmad SeyedAlinaghi, Mohammad Heydari, Amirali Karimi, Abdollah Mahdavi, Mehrnaz Mashoufi, Arezoo Sarmad, Peyman Mirghaderi, Ahmadreza Shamsabadi, Kowsar Qaderi, et~al. 2023.
\newblock Telemedicine technologies and applications in the era of covid-19 pandemic: A systematic review.
\newblock \emph{Health informatics journal}, 29(2):14604582231167431.

\bibitem[{Taira et~al.(2021)Taira, Kreger, Orue, and Diamond}]{taira2021pragmatic}
Breena~R Taira, Vanessa Kreger, Aristides Orue, and Lisa~C Diamond. 2021.
\newblock A pragmatic assessment of google translate for emergency department instructions.
\newblock \emph{Journal of General Internal Medicine}, 36(11):3361--3365.

\bibitem[{wai Yim et~al.(2024{\natexlab{a}})wai Yim, {Ben Abacha}, Fu, Sun, Xia, Yetisgen, and Krallinger}]{mediqa-m3g-task}
Wen wai Yim, Asma {Ben Abacha}, Velvin Fu, Zhaoyi Sun, Fei Xia, Meliha Yetisgen, and Martin Krallinger. 2024{\natexlab{a}}.
\newblock Overview of the mediqa-m3g 2024 shared task on multilingual and multimodal medical answer generation.
\newblock In \emph{Proceedings of the 6th Clinical Natural Language Processing Workshop}, Mexico City, Mexico. Association for Computational Linguistics.

\bibitem[{wai Yim et~al.(2024{\natexlab{b}})wai Yim, Fu, Sun, {Ben Abacha}, Yetisgen, and Xia}]{mediqa-m3g-dataset}
Wen wai Yim, Velvin Fu, Zhaoyi Sun, Asma {Ben Abacha}, Meliha Yetisgen, and Fei Xia. 2024{\natexlab{b}}.
\newblock Dermavqa: A multilingual visual question answering dataset for dermatology.
\newblock \emph{CoRR}.

\bibitem[{Yin et~al.(2022)Yin, Vahdat, Alvarez, Mallya, Kautz, and Molchanov}]{yin2022vit}
Hongxu Yin, Arash Vahdat, Jose~M Alvarez, Arun Mallya, Jan Kautz, and Pavlo Molchanov. 2022.
\newblock A-vit: Adaptive tokens for efficient vision transformer.
\newblock In \emph{Proceedings of the IEEE/CVF Conference on Computer Vision and Pattern Recognition}, pages 10809--10818.

\bibitem[{Zheng et~al.(2021)Zheng, Wang, Du, and Lu}]{zheng2021mutual}
Xiangtao Zheng, Binqiang Wang, Xingqian Du, and Xiaoqiang Lu. 2021.
\newblock Mutual attention inception network for remote sensing visual question answering.
\newblock \emph{IEEE Transactions on Geoscience and Remote Sensing}, 60:1--14.

\bibitem[{Zhou et~al.(2021)Zhou, Song, Zhou, Zhang, and Xing}]{zhou2021covid}
Changjian Zhou, Jia Song, Sihan Zhou, Zhiyao Zhang, and Jinge Xing. 2021.
\newblock Covid-19 detection based on image regrouping and resnet-svm using chest x-ray images.
\newblock \emph{Ieee Access}, 9:81902--81912.

\end{thebibliography}

\end{document}